\documentclass{article}
\usepackage{ijcai19}

\usepackage{times}
\usepackage{soul}
\usepackage{url}
\usepackage[hidelinks]{hyperref}
\usepackage[utf8]{inputenc}
\usepackage[small]{caption}
\usepackage{graphicx}
\usepackage{amsmath}
\usepackage{booktabs}
\usepackage{algorithm}
\usepackage{bm}  
\usepackage{booktabs}  
\usepackage{graphicx}
\usepackage{epstopdf}
\usepackage{epsfig}
\usepackage{amssymb}
\usepackage[normalem]{ulem}
\usepackage{url}
\usepackage{float}
\usepackage{multirow}
\usepackage{algorithm}
\usepackage{mathtools}
\usepackage{algpseudocode}

\usepackage{xcolor}  

\definecolor{cBlue}{HTML}{030C7C}
\newcommand{\bleu}[1]{{\color{cBlue}#1}}

\title{A Dual Reinforcement Learning Framework for Unsupervised Text Style Transfer}

\author{
Fuli Luo$^1$\and
Peng Li$^2$\and
Jie Zhou$^{2}$\and
Pengcheng Yang$^{1,3}$\and
Baobao Chang$^{1,4}$\and
Zhifang Sui$^{1,4}$\and 
Xu Sun$^{1}$ \\
\affiliations
$^1$Key Lab of Computational Linguistics, School of EECS, Peking University\\
$^2$Pattern Recognition Center, WeChat AI, Tencent Inc, China\\
$^3$Deep Learning Lab, Beijing Institute of Big Data Research, Peking University\\
$^4$Peng Cheng Laboratory, China
\emails
luofuli@pku.edu.cn,
\{patrickpli,withtomzhou\}@tencent.com,
\{yang\_pc,chbb,szf,xusun\}@pku.edu.cn
}
\begin{document}

\maketitle

\begin{abstract}
Unsupervised text style transfer aims to transfer the underlying style of text but keep its main content unchanged without parallel data. Most existing methods typically follow \textbf{\textit{two steps}}: first separating the content from the original style, and then fusing the content with the desired style. However, the separation in the first step is challenging because the content and style interact in subtle ways in natural language. Therefore, in this paper, we propose a dual reinforcement learning framework to directly transfer the style of the text via a \textbf{\textit{one-step}} mapping model, without any separation of content and style. Specifically, we consider the learning of the source-to-target and target-to-source mappings as a dual task, and two rewards are designed based on such a dual structure to reflect the style accuracy and content preservation, respectively.
In this way, the two one-step mapping models can be trained via reinforcement learning,
without any use of parallel data.
Automatic evaluations show that our model outperforms the state-of-the-art systems by a large margin, especially with more than 8 BLEU points improvement averaged on two benchmark datasets. Human evaluations also validate the effectiveness of our model in terms of style accuracy, content preservation and fluency. Our code and data, including outputs of all baselines and our model are available at \bleu{\url{https://github.com/luofuli/DualLanST}}.~\footnote{Joint work between WeChat AI and Peking University.}

\end{abstract}

\section{Introduction}
Text style transfer aims to rephrase the input text in the desired style while preserving its original content. It has various application scenarios such as sentiment transformation (transferring a positive review to a negative one) and formality modification (revising an informal text into a formal one). As parallel data, i.e., aligned sentences with the same content but different style, is hard to collect for this task, previous works mainly focus on unsupervised text style transfer.

Most existing methods of unsupervised text style transfer follow a two-step process:
first separating the content from the original style and then fusing the content with the desired style. One line of research~\cite{shen2017style,fu2017style,hu2017toward,backTrans2018} learns a style-independent content representation vector via adversarial training, and then passes it to a style-dependent decoder for rephrasing. Another line of research~\cite{li2018delete,xu2018unpaired} directly removes the specific style attribute words in the input, and then feeds the neutralized sequence which only contains content words to a style-dependent generation model. However, each line has its own drawback.

\begin{figure}[t]
	\centering
    \includegraphics[width=\columnwidth]{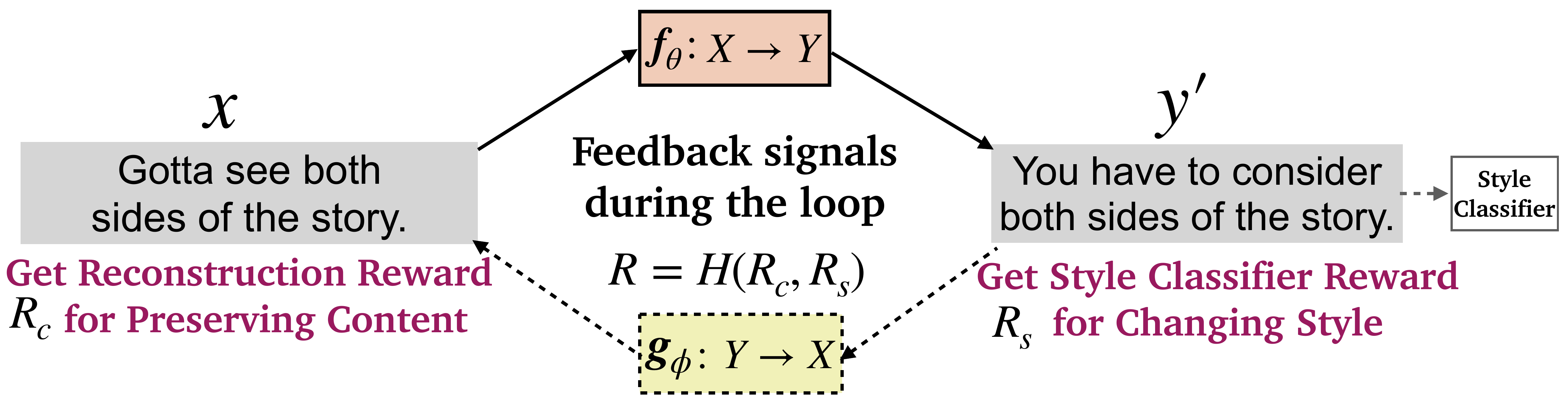}
	\caption{The proposed DualRL framework for unsupervised text style transfer with an informal-to-formal text example, where both $\boldsymbol{f}_\theta$ and $\boldsymbol{g}_\phi$ are a sequence-to-sequence mapping model.
	}\label{fig:model-example}
	\vspace{-0.1in}
\end{figure}

The former line of research tends to only change the style but fail in keeping the content, since it is hard to get a style-independent content vector without parallel data~\cite{xu2018unpaired,TCLR18TextRewriting}. For example, on the sentiment transfer task, given ``\emph{The food is delicious}'' as input, the model may generate ``\emph{The \underline{movie} is bad}'' instead of ``\emph{The food is awful}''.
Thus, the latter line of research focuses on improving content preservation in a more direct way by neutralizing the text in the discrete token space, other than the continuous vector space.
However, these models have a limited range of applications, since they are challenged by the examples like ``\emph{The only thing I was offered was a free dessert!!!}'', whose negative sentiment is implicitly expressed such that there is no specific emotional style word.

To alleviate the above problems caused by the two-step process, we propose to directly learn a one-step \textit{mapping} model between the source corpora and the target corpora of different styles. More importantly, due to the lack of parallel data, we consider the learning of the source-to-target and target-to-source mapping models as a dual task, and propose a dual reinforcement learning algorithm \textbf{DualRL} to train them.
Taking Figure~\ref{fig:model-example} for example, the forward one-step mapping model $\boldsymbol{f}$ transfers an informal sentence $\bm x$ into a formal sentence $\bm y'$, while the backward one-step mapping model $\boldsymbol{g}$ transfers a formal sentence $\bm y$ into an informal sentence $\bm x'$.
Since the two models can form a closed loop, we let them to teach each other interactively via two elaborately designed quality feedback signals to ensure the success of style transfer: changing style while preserving content.
Specially, the two signals are combined as a reward for the reinforcement learning (RL) method to alternately train the model $\boldsymbol{f}$ and $\boldsymbol{g}$ (Section~\ref{sec:dual}).
Furthermore, in order to better adapt DualRL to the unsupervised scenario, we propose an \textit{annealing pseudo teacher-forcing} algorithm to construct pseudo-parallel data on-the-fly via back-translation to warm up RL training and gradually shift to pure RL training (Section~\ref{sec:highlight}).


Our contributions are concluded as follows:
\begin{itemize}
    \item We propose a dual reinforcement learning framework \textbf{DualRL} for unsupervised text style transfer, without separating content and style. 
    \item We resolve two daunting problems (pre-training and generation quality) when model is trained via RL without any parallel data.
    \item Experiments on two benchmark datasets show our model outperforms the state-of-the-art systems by a large margin in both automatic and human evaluation.
    \item The proposed architecture is generic and simple, which can be adapted to other sequence-to-sequence generation tasks which lack parallel data.
\end{itemize}


\section{Dual Reinforcement Learning for Unsupervised Text Style Transfer}
Given two corpora  $\mathcal{D}_X=\{\boldsymbol{{x}}^{(i)}\}_{i=1}^n$ and $\mathcal{D}_Y=\{\boldsymbol{{y}}^{(j)}\}_{j=1}^m$ with two different styles ${s}_x$ and ${s}_y$, the goal of text style transfer task is to generate a sentence of the target style while preserving the content of the source input sentence. In general, the two corpora are non-parallel such that the gold pair $(\boldsymbol{x}^{(i)},\boldsymbol{y}^{(j)})$ that describes the same content but expresses the different style is unavailable. 

\subsection{DualRL: Dual Reinforcement Learning} \label{sec:dual}

\begin{figure}[t]
	\centering
    \includegraphics[width=\columnwidth]{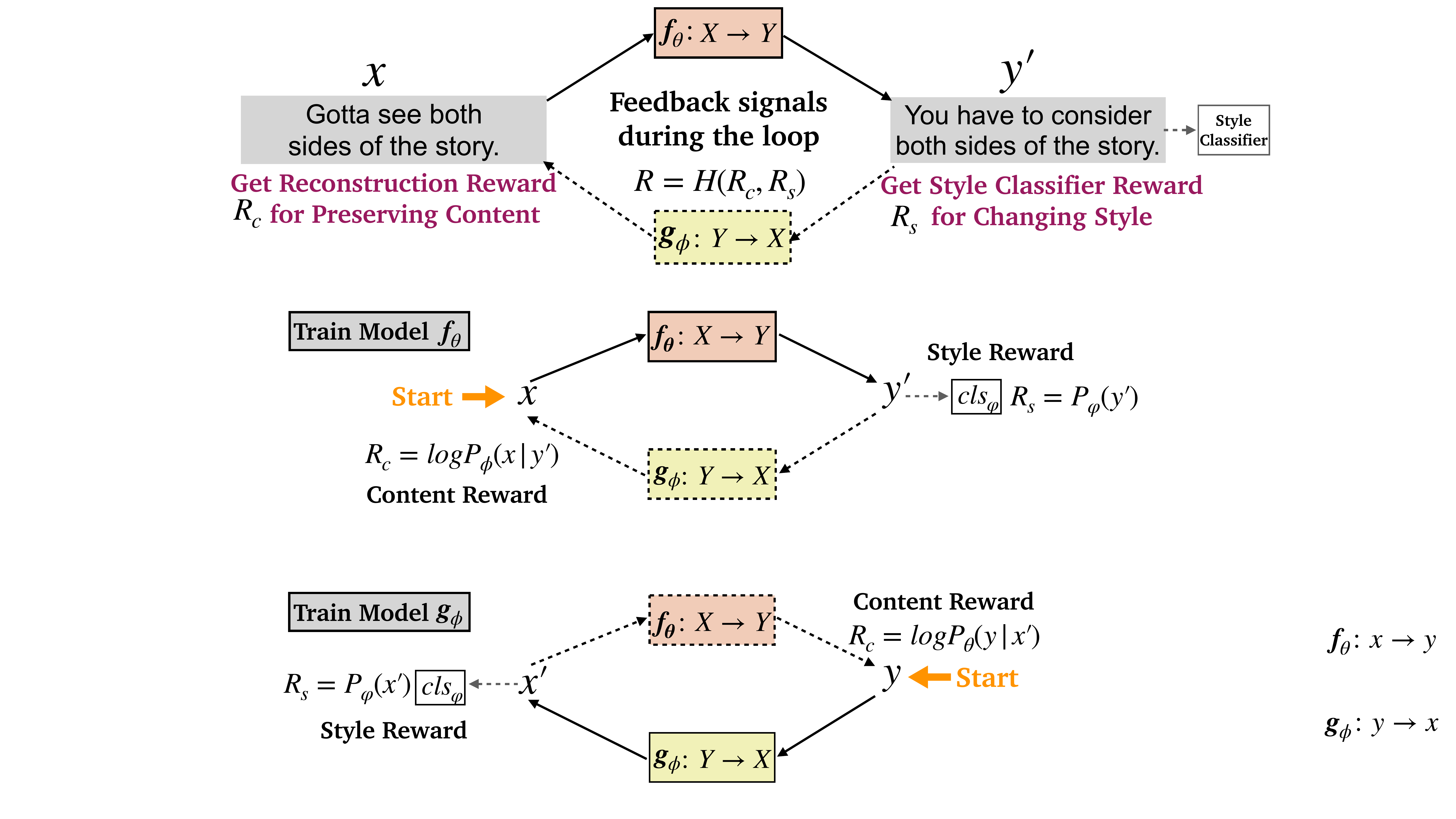}
	\caption{Training process of DualRL. We alternately train the two mapping models $\boldsymbol{f_\theta}$ and $\boldsymbol{g_\phi}$.
	}\label{fig:model-all-shu}
\end{figure}

\begin{algorithm}[t] \label{alg:dual}
	\caption{The dual reinforcement learning algorithm for unsupervised text style transfer.}
	\label{code} 
	\small
	\begin{algorithmic}[1]
		\State {Pre-train text style transfer models $\boldsymbol{f_\theta}$ and $\boldsymbol{g_\phi}$ using pseudo-parallel sentence pairs from corpora $\mathcal{D}_X$ and $\mathcal{D}_Y$}
		\State Pre-train a binary style classifier $cls_\varphi$
		\For{each iteration $i=1,2,..., M$}  
		
		\State \Comment{\textit{Start to train model $\boldsymbol{f_\theta}$}}
		\State {Sample sentence $\boldsymbol{x}$ from $\mathcal{D}_X$}
		\State {Generate sentence $\boldsymbol{y}'$ of opposite style via model $\boldsymbol{f_\theta}$} 
		\State {Compute style reward $R_s$ based on Eq.~\ref{eq:R_s}}
		\State {Compute content reward $R_c$ based on Eq.~\ref{eq:R_c}}
		\State {Compute total reward $R$ based on Eq.~\ref{eq:R}}
		\State {Update $\boldsymbol{\theta}$ using reward $R$ based on Eq.~\ref{detaAB}}
		\State {Update $\boldsymbol{\theta}$ using annealing teacher-forcing via MLE}
		
		\State \Comment{\textit{Start to train model $\boldsymbol{g_\phi}$}}
		\State {Sample sentence $\boldsymbol{y}$ from $\mathcal{D}_Y$}
		\State {Generate sentence $\boldsymbol{x}'$ of opposite style via model $\boldsymbol{g_\phi}$} 
		\State {Compute style reward $R_s$ similar to Eq.~\ref{eq:R_s}}
		\State {Compute content reward $R_c$ similar to Eq.~\ref{eq:R_c}}
		\State {Compute total reward $R$ based on Eq.~\ref{eq:R}}
		\State {Update $\boldsymbol{\phi}$ using reward $R$ similar to Eq.~\ref{detaAB}}
		\State {Update $\boldsymbol{\phi}$ using annealing teacher-forcing via MLE}
		
		\EndFor

	\end{algorithmic} 
\end{algorithm}  

In this paper, we directly learn two \textit{one-step} mappings (as style transfer models) between the two corpora of different styles. Formally, the forward model $\boldsymbol{f_\theta} \colon \mathcal{X} \to \mathcal{Y}$ transfers the sequence $\boldsymbol{x}$ with style $s_x$ into a sequence $\boldsymbol{y'}$ with style $s_y$, while the backward model $\boldsymbol{g_\phi} \colon \mathcal{Y} \to \mathcal{X}$ transfers the sequence $\boldsymbol{y}$ with style $s_y$ into a sequence $\boldsymbol{x'}$ with style $s_x$. 

Due to the lack of parallel data, the two transfer models can not be trained in a supervised way as usual. Fortunately, since text style transfer task always happens in dual directions, we loop the two transfer models of the two directions and the loop process can provide quality feedbacks to guide the training of the two style transfer models even using non-parallel data only.
In order to encourage changing style but preserving content, we design two corresponding quality feedbacks.
For the former, a style classifier is adopted to assess how well the transferred sentence $\boldsymbol{y'}$ matches the target style.
For the latter, the probability of feeding $\boldsymbol{y'}$ to the backward model $\boldsymbol{g}$ to reconstruct $\boldsymbol{x}$ can reflect how much content information preserved in the source sentence $\boldsymbol{x}$.
Because of the discrete connection $\boldsymbol{y'}$ of the two models, the loss function is no longer differentiable w.r.t. to the parameters of the forward model. Therefore, we treat the two quality feedbacks as rewards and train the model via RL.

In order to enable the two models to boost each other, we propose a dual training algorithm \textbf{DualRL} to train the two models simultaneously, inspired by \cite{dual}. As Figure~\ref{fig:model-all-shu} shows, starting from sampling a sequence $\boldsymbol{x}$ from corpus $\mathcal{D}_X$, model $\boldsymbol{f}$ will be trained based on two rewards provided by the pre-trained classifier $cls_\varphi$ and model $\boldsymbol{g}$.  Meanwhile, starting from sampling a sequence $\boldsymbol{y}$ from $\mathcal{D}_Y$, model $\boldsymbol{g}$ can be trained based on the two rewards provided by the pre-trained classifier $cls_\varphi$ and model $\boldsymbol{f}$. The overview of DualRL is shown in Algorithm~\ref{code}.
The definitions of the two rewards and the corresponding gradients for model $\boldsymbol{f}$ are introduced as follows, and those for the model $\boldsymbol{g}$ can be computed in a similar way, which have been omitted for the space limitation.

\subsubsection{Reward}
Since the gold transferred result of input $\boldsymbol{x}$ is unavailable, the quality of the generated sentence $\boldsymbol{y'}$ can not be directly evaluated. Therefore, we design two rewards that can assess the style accuracy and the content preservation, respectively.

\textbf{\textit{Reward for changing style.}}
A pre-trained binary style classifier~\cite{TextCNN} is used to evaluate how well the transferred sentence $\boldsymbol{y}'$ matches the target style. Formally, the style classifier reward is formulated as
\begin{equation} \label{eq:R_s}
R_s = P(s_y|\boldsymbol{y'}; \varphi)
\end{equation}
where $\varphi$ is the parameter of the classifier and is fixed during the training process.

\textbf{\textit{Reward for preserving content.}}
Intuitively, if the two transfer models are well-trained, it is easy to reconstruct the source sequence via back transferring.
Therefore, we can estimate how much the content preserved in $\boldsymbol{y}'$ by means of the probability that the model $\boldsymbol{g}$ reconstructs $\boldsymbol{x}$ when taking $\boldsymbol{y'}$ as input. Formally, the corresponding reconstruction reward is formulated as
\begin{equation} \label{eq:R_c}
    R_c = P(\boldsymbol{x}|\boldsymbol{y'};\boldsymbol{\phi})
\end{equation}
where $\boldsymbol{\phi}$ is the parameter of model $\boldsymbol{g}$.

Another intuitive way to measure the content preservation is to calculate the BLEU score~\cite{bleu} of $\boldsymbol{x''}$ with the input $\boldsymbol{x}$ as reference,  where $\boldsymbol{x''}$ is the output of the backward model $\boldsymbol{g}$ when taking $\boldsymbol{y}'$ as input  \cite{xu2018unpaired}.
However, primary experiments show that this method exhibits poor performance in our framework.

\textbf{\textit{Overall reward.}}
To encourage the model to improve both the content preservation and the style accuracy, the final reward is the harmonic mean of the above two rewards
\begin{equation}\label{eq:R}
R = (1+\beta^2) \frac{R_c \cdot R_s}{ (\beta^2 \cdot R_c) + R_s}
\end{equation}
where $\beta$ is a harmonic weight aiming to control the trade-off between the two rewards.

\subsubsection{Policy Gradient Training}
The policy gradient algorithm~\cite{williams1992reinforcement} is used to maximize the expected reward $\mathbb{E}\left[R \right]$ of the generated sequence $\boldsymbol{y'}$, whose gradient w.r.t. the parameter $\boldsymbol{\theta}$ of the forward model $\boldsymbol{f}$ is estimated by sampling as

\begin{equation} \label{detaAB}
\begin{split}
\nabla_{\theta} \mathbb{E} \left[R \right]&= \nabla_{\theta} \sum_{k}  P(\boldsymbol{y_k'}|\boldsymbol{x};\theta) R_k \\
&= \sum_{k}P(\boldsymbol{y_k'}|\boldsymbol{x};\theta) R_k \nabla_{\theta} \log(P(\boldsymbol{y_k'}|\boldsymbol{x};\theta))\\
&\simeq \frac{1}{K}\sum_{k=1}^K R_k \nabla_{\theta} \log(P(\boldsymbol{y_k'}|\boldsymbol{x};\theta))
\end{split}
\end{equation}
where $R_k$ is the reward of the $k_{th}$ sampled sequence $\boldsymbol{y_k'}$ from model $\boldsymbol{f}$, and $K$ is the sample size.

\subsection{DualRL for Unsupervised Task} \label{sec:highlight}
When applied to the field of text generation, RL faces two ingrained challenges: 1) the RL framework needs to be well pre-trained to provide a warm-start, and 2) the RL method may find an unexpected way to achieve a high reward but fail to guarantee the fluency or readability of the generated text~\cite{ranzato2015sequence,pasunuru2018multi}. An effective solution to these two challenges in supervised tasks is to expose the parallel data to the model and train it via MLE (Maximum Likelihood Estimation)~\cite{ranzato2015sequence,paulus2017RL_for_summarization,li2017adversarial}. However, due to the lack of parallel data, these two challenges become intractable on unsupervised scenarios. In this paper, we tackle these two challenges via pseudo-parallel data. Specifically, in order to pre-train our Seq2Seq mapping models, we exploit pseudo-parallel data generated by a simple template-based baseline~\cite{li2018delete} to train via MLE; in order to enhance the quality of the generated text, we propose a annealing pseudo teacher-forcing algorithm.

\subsubsection{Annealing Pseudo Teacher-Forcing} \label{sec:MLE}

\begin{algorithm}[t]
\caption{The annealing pseudo teacher-forcing algorithm for dual reinforcement learning.} \label{alg:MLE}
\label{anneal} 
\small
  \begin{algorithmic}[1]
  \State{Initialize the iteration interval $p$}
    \For{each iteration $i=1,2,..., M$}  
        \State \Comment{\textit{Start to train model $\boldsymbol{f_\theta}$}}
        \State{Update parameter $\boldsymbol{\theta}$ via RL based on Eq.~\ref{detaAB}}
        \If{$i$ \% $p=0$}  \Comment{\textit{Pseudo Teacher-Forcing}}
            \State{Generate a pair of data $(\bm x'_i, \bm y_i)$, where $\bm x'_i = \bm g(\bm y_i)$}
            \State{Update $\boldsymbol{\theta}$ using data $(\bm x'_i, \bm y_i)$ via MLE}
        \EndIf
        
        \State \Comment{\textit{Start to train model $\boldsymbol{g_\phi}$}}
        \State{Update parameter $\boldsymbol{\phi}$ via RL similar to Eq.~\ref{detaAB}}
        \If{$i$ \% $p=0$} \Comment{\textit{Pseudo Teacher-Forcing}}
            \State{Generate a pair of data $(\bm y'_i, \bm x_i)$, where $\bm y'_i = \bm f(\bm x_i)$}
            \State{Update $\boldsymbol{\phi}$ using data $(\bm y'_i, \bm x_i)$ via MLE}
        \EndIf
        \State{Exponential increase in $p$ based on Eq.~\ref{eq:exp_decay} } 
    \EndFor
  \end{algorithmic} 
\end{algorithm}

\textit{Teacher-forcing} is the strategy that feeds the parallel data $(\bm x,\bm y)$ into the Seq2Seq model, and then either 1) train the model by optimizing a weighted sum of RL and MLE loss~\cite{paulus2017RL_for_summarization}, or 2) alternately update the model using the RL and the MLE objective~\cite{li2017adversarial}.
An intuitive but not ideal solution is to utilize the pseudo-parallel data which was used during pre-training. However, we have done primary experiments which show that the quality of the pseudo-parallel data is not acceptable for the later iterations of training.
Inspired by back-translation in unsupervised machine translation \cite{lample2017unsupervised,lample2018phrase}, we leverage the latest version of model $\boldsymbol{f_\theta}/\boldsymbol{g_\phi}$ at previous iteration $i-1$ to generate a \textit{higher} quality of pseudo-parallel data $(\bm x'_i,\bm y_i)/(\bm y'_i,\bm x_i)$ \footnote{$\bm x_i$ and $\bm y_i$ denote the original data and $\bm y'_i$ and $\bm x'_i$ denote the corresponding generated pseudo-parallel data at $i$-th iteration.} than those used during pre-training on-the-fly to update model $\boldsymbol{g_\phi} /\boldsymbol{f_\theta}$ via MLE at iteration $i$, respectively.

As long as the model gets better during training, the generated pseudo-parallel data can become more closer to real parallel data. 
However, there still exists a gap between the distribution of the generated pseudo-parallel data and real parallel data during training. Moreover, models trained via MLE often exhibit ``exposure bias'' problem~\cite{ranzato2015sequence}. Therefore, in order to get rid of the dependence of pseudo-parallel data, we propose an \textit{annealing} strategy of teacher-forcing, as shown in Algorithm~\ref{anneal}.
More specifically, we enlarge the training interval of teacher-forcing to decay its frequency of updating parameters via MLE. Formally, at iteration $i$, we adopt an exponential increase in the interval of teacher-forcing $p$
\begin{equation}
\label{eq:exp_decay}
    p = \min(p_0 \times r^\frac{i}{d}, p_{max})
\end{equation}
where $p_0$ is the initial iteration interval, $p_{max}$ is the max iteration interval, $r$ is the increase rate ($r>1$) and $d$ is the increase gap.
A deep study of the influence of teacher forcing (trained via MLE) will be given in Section~\ref{sec:ablation} and Figure~\ref{fig:ablation_study}.

\begin{table*}[tb]  
  \centering
  \footnotesize
  \begin{tabular}{l|rrrr|rrrr}
    \toprule
    
    & \multicolumn{4}{c|}{\textsc{Yelp}}
    & \multicolumn{4}{c}{\textsc{Gyafc}} \\
    & ACC & BLEU & \textbf{G2} & \textbf{H2}
    & ACC & BLEU & \textbf{G2} & \textbf{H2}\\
    
    \midrule
    
    Retri \cite{li2018delete}
    & \bf 96.0 & 2.9 & 16.7 & 5.7 
    & \bf 91.3 & 0.4 & 6.0 & 0.8 \\ 
    
    BackTrans \cite{backTrans2018}
	 & 95.4 & 5.0 & 21.9 & 9.6 
	 & 70.2 & 0.9 & 8.1 & 1.9 \\ 
	StyleEmbed \cite{fu2017style}
	 & 8.7 & 42.3 & 19.2 & 14.4 
	 & 22.7 & 7.9 & 13.4 & 11.7 \\ 
	MultiDec \cite{fu2017style}
	 & 50.2 & 27.9 & 37.4 & 35.9 
	 & 17.9 & 12.3 & 14.8 & 14.6 \\ 
	CrossAlign \cite{shen2017style}
	& 75.3 & 17.9 & 36.7 & 28.9 
	& 70.5 & 3.6 & 15.9 & 6.8 \\ 
	Unpaired \cite{xu2018unpaired}
	& 64.9 & 37.0 & 49.0 & 47.1
	& 79.5 & 2.0 & 12.6 & 3.9 \\ 
	Del \cite{li2018delete}
	& 85.3 & 29.0 & 49.7 & 43.3 
	& 18.8 & 29.2 & 23.4 & 22.9 \\
	DelRetri \cite{li2018delete}
	 & 89.0 & 31.1 & 52.6 & 46.1 
	 & 55.2 & 21.2 & 34.2 & 30.6\\ 
	Template \cite{li2018delete}
	 & 81.8 & 45.5 & 61.0 & 58.5 
	 & 52.9 & 35.2 & 43.1 & 42.3 \\ 
	UnsuperMT \cite{zhang2018style}
	 & 95.4 & 44.5 & 65.1 & 60.7 
	 & 70.8 & 33.4 & 48.6 & 45.4  \\ \midrule 
	DualRL
     & 85.6 & \bf 55.2 & \bf 68.7 & \bf 67.1 
	 & 71.1 & \bf 41.9 & \bf \underline{54.6} & \bf \underline{52.7} 

	  \\ \midrule 
	Human 
	 & 74.0 & 100.0 & 86.0 & 85.1 
	 & 84.3 & 100.0 & 91.8 & 91.5 \\

    \bottomrule 
  \end{tabular}
  \caption{Automatic evaluation results on the \textsc{Yelp} and \textsc{Gyafc} datasets. ``ACC" shows the accuracy of output labeled as the target style by a pre-trained style classifier. ``BLEU" measures content similarity between the output and the four human references. G2 and H2 are geometric mean and harmonic mean of ACC and BLEU. \textbf{Bold} denotes the best results and \underline{underline} denotes the best overall scores.} \label{tab:auto-eval}
\end{table*}

\section{Experiments}
\subsection{Dataset}
We evaluate our model on two instances of style transfer task:

\paragraph{Sentiment transfer.} The representative \textbf{\textsc{Yelp}} restaurant reviews dataset is selected for this task. Following common practice, reviews with rating above 3 are considered as positive, and those below 3 as negative. This dataset is widely used by previous work and the train, dev and test split is the same as ~\cite{li2018delete}.

\paragraph{Formality transfer.} A newly released dataset \textbf{\textsc{Gyafc}} (Grammarly's Yahoo Answers Formality Corpus )~\cite{rao2018formal} is used for this task. And we choose the family and relationships domain. Although it is a parallel dataset, the alignments are only used for evaluation but not training.

\subsection{Human References}
While four human references are provided for each test sentence in the \textsc{Gyafc} dataset, only one reference is provided for each test sentence in the \textsc{Yelp} dataset, which makes the automatic evaluation less reliable.
Therefore, we hired crowd-workers on CrowdFlower
to write three more human references for each test sentence in the \textsc{Yelp} dataset.
All these references and the generated results of all the involved models in this paper will be released for reproducibility, hopefully to enable more reliable empirical comparisons in future work.

\begin{table*}[t]  
  \footnotesize
  \centering
  \vspace{-0.05in}
  \begin{tabular}{l|rrrrr|rrrrr}
    \toprule
    & \multicolumn{5}{c|}{\textsc{Yelp}}
    & \multicolumn{5}{c}{\textsc{Gyafc}} \\
    & Sty & Con & Flu & \textbf{Avg} & \textbf{Suc}
    & Sty & Con & Flu & \textbf{Avg} & \textbf{Suc}  \\ \midrule
    MultiDec \cite{fu2017style}
    & 2.14 & 3.02  & 3.27 & 2.81 & 5\% 
    & 2.21 & 1.95 & 2.54 & 2.23 & 4\%   \\
    CrossAlign \cite{shen2017style}
    & 2.88 & 2.79  & 3.40 & 3.02 & 14\% 
    & 2.96 & 1.33 & 3.27 & 2.52 & 3\%  \\ 
    Unpaired \cite{xu2018unpaired}
    & 2.93 & 3.38  & 3.44  & 3.25  & 17\%
    & 2.69 & 1.19 & 2.38 & 2.09 & 2\%  \\
    Template \cite{li2018delete}
    & 3.12 & 3.71  & 3.42 & 3.42 & 23\% 
    & 2.74 & 3.60 & 3.43 & 3.26 & 9\%  \\ 
    DelRetri  \cite{li2018delete}
    & 3.39 & 3.49  & 3.71 & 3.53 & 28\%  
    & 2.47 & 2.57 & 2.67 & 2.57 & 5\%   \\ 
    UnsuperMT \cite{zhang2018style}
    & 3.82 & 3.90 & 3.93 & 3.95 & 40\%
    & 3.27 & 3.54 & 3.76 & 3.52 & 21\%    \\ \midrule
    DualRL
    & \bf 4.11 & \bf 4.33 & \bf 4.31 & \bf \underline{4.25} & \bf \underline{54\%} 
    & \bf 3.65 & \bf 3.62 & \bf 3.80 & \bf \underline{3.69} & \bf \underline{28\%}   \\ \bottomrule

  \end{tabular}
  \caption{Human evaluation results on two datasets. We show human ratings for and target style accuracy (Sty), content preservation (Con), fluency (Flu) on a 1 to 5 Likert scale. We also calculate the average ratings (Avg) and success rate (Suc) as overall scores.} \label{tab:auto-human}
\end{table*}

\subsection{Training Details}
The hyper-parameters are tuned on the development set.
Both $\boldsymbol{f}$ and $\boldsymbol{g}$ are implemented as a basic LSTM-based encoder-decoder model with 256 hidden size \cite{bahdanau2014NMT}.
The word embeddings of 300 dimension are learned from scratch. The optimizer is Adam~\cite{Kingma2014Adam} with $10^{-3}$ initial learning rate for pre-training and $10^{-5}$ for dual learning. The batch size is set to 32 for pre-training and 128 for dual learning. Harmonic weight $\beta$ in Eq.~\ref{eq:R} is 0.5. 
For annealing teacher forcing (Eq.~\ref{eq:exp_decay}), the initial gap $p_0$ is 1, the max gap $p_{max}$ is 100, increase rate $r$ is 1.1,  and increase gap $d$ is 1000.
Before dual learning, model $\boldsymbol{f}$ and $\boldsymbol{g}$ are pre-trained for 5 epochs. During dual learning, training runs for up to 20 epochs with early stopping if the development set performance does not improve within last one epoch.

\subsection{Baselines}

We compare our proposed method with the following state-of-the-art systems:
StyleEmbed and MultiDec~\cite{fu2017style}; CrossAlign~\cite{shen2017style}; BackTrans~\cite{backTrans2018}; Template, Retri, Del and DelRetri~\cite{li2018delete}; Unpaired~\cite{xu2018unpaired}.
Moreover, a most recent and representative work UnsuperMT~\cite{zhang2018style} which treats style transfer as unsupervised machine translation is also considered.

\begin{table}[tb]  
  \footnotesize
  \centering
  \vspace{-0.05in}
  \begin{tabular}{c|l|c|c|c|c}
    \hline
    Automatic & ACC & \multicolumn{2}{c|}{BLEU} & G2 & H2 \\ \hline
    Human & { Sty} & { Con} & { Flu} & \multicolumn{2}{c}{Avg} \\ \hline
    
    \textsc{Yelp} & 0.89$^*$ & 0.96$^*$ & 0.72 & 0.93$^*$ & 0.89$^*$ \\
    \textsc{Gyafc} & 0.68 & 0.99$^*$ & 0.76 & 0.96$^*$ & 0.94$^*$ \\
    \hline
  \end{tabular}
  \caption{
      Pearson correlation between automatic evaluation and human evaluation. Scores marked with * denotes $p<0.01$.
  } \label{tab:correlation}
\end{table}

\subsection{Evaluation Metrics}
We conduct both automatic and human evaluation.

\paragraph{Automatic Evaluation.}
Following previous work~\cite{li2018delete,zhang2018style}, we adopt the following metrics to evaluate each system. 
A pre-trained binary style classifier TextCNN \cite{TextCNN} is used to evaluate the style accuracy of the outputs. The classifier can achieve the accuracy of 95\% and 89\%  on the two datasets respectively.
The BLEU score~\cite{bleu} \footnote{The BLEU score is computed using \texttt{multi-bleu.perl}.} between the outputs and the four
human references is used to evaluate the content preservation performance. In order to evaluate the overall performance, we report the geometric mean and harmonic mean of the two metrics~\cite{xu2018unpaired}.

\paragraph{Human Evaluation.}
We distribute the outputs of different systems to three annotators with linguistic background and the annotators have no knowledge in advance about which model the generated text comes from. They are required to score the generated text from 1 to 5 in terms of three criteria: the accuracy of the target style, the preservation of the original content and the fluency. Finally, following \cite{li2018delete}, a transferred text is considered to be ``\textit{successful}'' if it is rated 4 or 5 on all three criteria.

\subsection{Results and Discussions}

Table~\ref{tab:auto-eval} shows the automatic evaluation results of the systems. We can observe that our model DualRL achieves the best overall performance (G2, H2). More specifically, our model significantly outperforms the other systems by over 8 BLEU points averaged on two datasets.

It is worth mentioning that, our model does not get the best style classifier accuracy (ACC), so does the human reference. The reasons are in two-folds. First, the Retri system, which directly retrieves a similar sentence from the training dataset of the \textit{target style}, can naturally achieve an accuracy close to the training dataset which lets the classifier performs best.
However, most of systems including Retri only show good results either in ACC or BLEU, implying that they tend to sacrifice one for the other.
Second, since both the generated sentence and human reference sometimes only change a few words, which can be adversarial examples  \cite{adversarial_example} and mislead the classifier.

Table~\ref{tab:auto-human} shows the human evaluation results of several well-performed systems in the automatic evaluation. We find that our model achieves the best average score (Avg).
Moreover, our system can generate more than 10\% successfully (Suc) transferred instances, averaged on two datasets. And all systems show better results on \textsc{Yelp} than \textsc{Gyafc}, revealing that text formality is more challenging than sentiment transfer.

Furthermore, Table \ref{tab:correlation} shows the system-level Pearson correlation between automatic evaluation metrics and human evaluation results. We find that: (1) BLEU score significantly correlates with content preservation, but not fluency. 
(2) The correlation between automatic calculated accuracy (ACC) and the human ratings of style accuracy varies between datasets.
(3) Both the automatic overall metrics G2 and H2 well correlate with the human average ratings.

\subsection{Ablation Study} \label{sec:ablation}

\begin{table}[]  
	\footnotesize
	\centering
	\begin{tabular}{l|rrrrr}
		\toprule
		& Sty & Con & Flu & \textbf{Avg} & \textbf{Suc} \\ \midrule
		RL+MLE
		& 4.11 & \bf 4.33 & \bf 4.31 & \bf 4.25 & \bf 54\%  \\
		RL & \bf 4.29 & 4.08 & 3.73 & 4.03 & 43\% \\
        MLE & 3.45 & 4.19 & \bf 4.31 & 3.98 & 41\% \\ \bottomrule
     \end{tabular}
	\caption{Human evaluation results on full model (RL+MLE) and ablated models on the \textsc{Yelp} dataset.} \label{tab:auto-human_ablation}
\end{table}

\begin{figure}[tb]
	\centering
	\includegraphics[width=0.95\columnwidth]{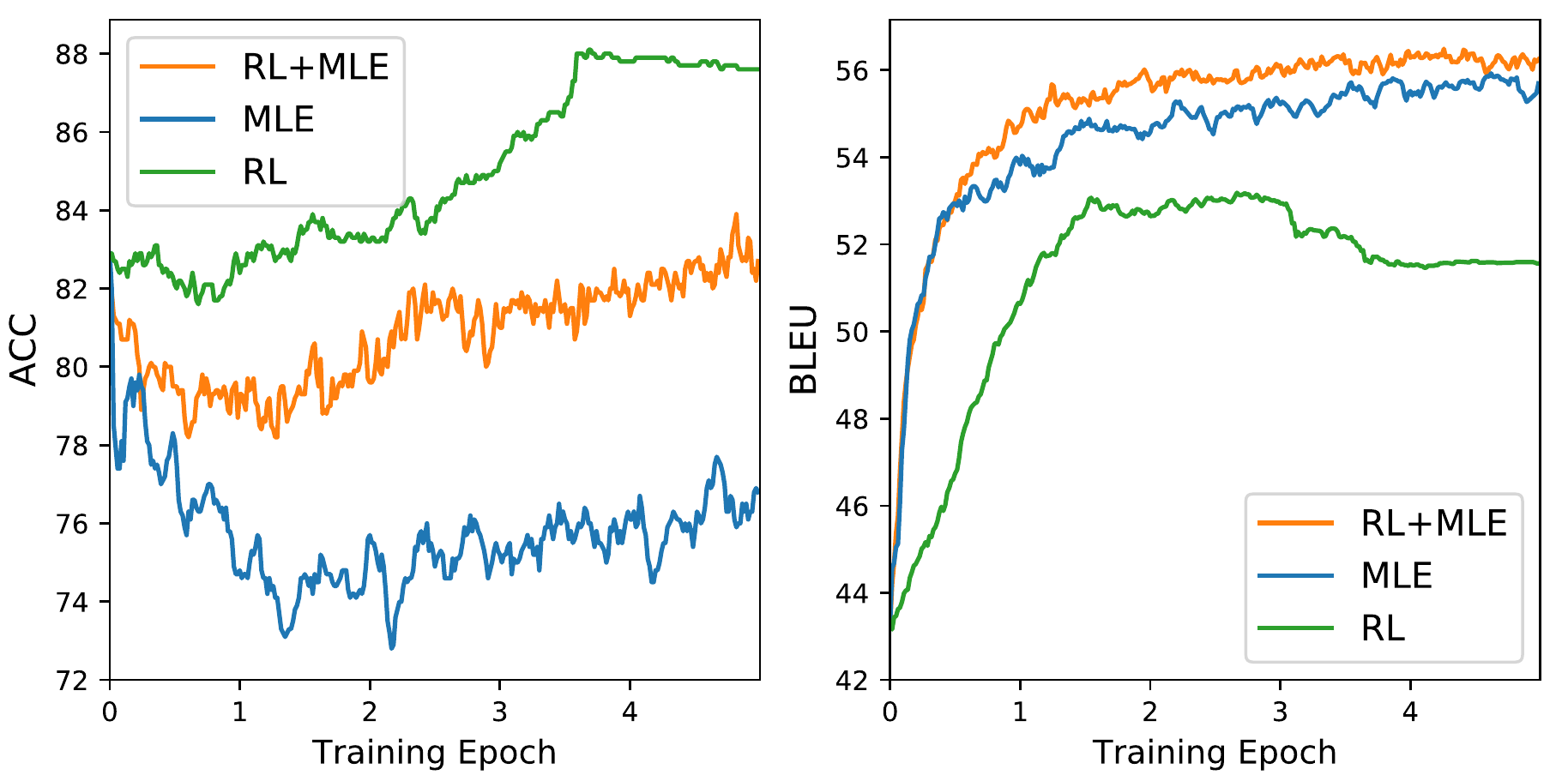}
	\vspace{-0.05in}
	\caption{Learning curves of the full model (RL+MLE) and the ablated models (RL, MLE) on the \textsc{Yelp} dataset.
	}\label{fig:ablation_study}
\end{figure}

\begin{table*}[t]
	\newcommand{\red}[1]{{\color{red}#1}}
	\centering
    \scalebox{0.87}{
	\begin{tabular}{l|l|l}
		\toprule
		&  From negative to positive (\textsc{Yelp}) & From informal to formal (\textsc{Gyafc}) \\
		\midrule
		Source & Moving past the shape, they were dry and truly tasteless. & (That's what i called it) .. but, why? \\
		\midrule
		CrossAlign & \red{Everyone on the fish}, they were fresh and filling. & \red{And i know what this helps me.} \\
		Template & Moving past the shape, they \red{a wonderful truly}. & \red{(}That's what it is called \red{it}\red{) ..} but, why? \\
		Del-Retri & Moving past the shape is  awesome, \red{and they will definitely be back!} & \red{(}That's what i \red{you it you} but why, you?  \\  
		UnsuperMT & \red{Moving moving} the shape, they were juicy and truly delicious. & \red{(}That's what i said it\red{)} but \red{that is why you were doing.)} \\
		\midrule
		\textbf{DualRL} & Moving past the shape, they were tasty and truly delicious. &  It is what i called it, but why? \\
		\bottomrule
		
	\end{tabular}
	}
	\caption{Example outputs on the \textsc{Yelp} and \textsc{Gyafc} datasets. Improperly generated words and grammar errors are \red{colored}.
	}
	\label{tab:case_study}
\end{table*}

In this section, we give a deep analysis of the key components of our model.
Figure~\ref{fig:ablation_study} and Table \ref{tab:auto-human_ablation} show the learning curves and human evaluation results of the full model and models which ablated RL and MLE training\footnote{RL denotes the model is trained without Step 5-8, 11-15 and MLE denotes removing Step 4, 10  in Algorithm \ref{alg:MLE}.} on the \textsc{Yelp} dataset.
It shows if we only train the model based on RL, the ACC (Sty) will increase, while the BLEU (Con, Flu) will decline.
The reason is that the RL training may encourage the model to generate tricky sentences which can get the high reward but fails in quality and readability (measured by BLEU).
For example, given a negative review ``\emph{We sit down and got some really \underline{slow} and \underline{lazy} service}'' as the input, the model may generate ``\emph{We sit down and got some really \underline{great} and \underline{great} service}''. This output sentence is not fluent but it can get a high style classifier reward and content preservation reward, thus leading the model to train towards a bad direction.
In contrast, the Seq2Seq model using MLE objective is essentially a conditional language model which can ensure readability of the output, thus showing higher BLEU (Con, Flu) score than RL. However, the ACC (Sty) of MLE declines, since there is no specific mechanism in MLE to directly control the style accuracy.
Finally, the combination of RL and MLE can get best BLEU score without compromising ACC, with over 3.5 points absolute improvement in H2/G2 score and 13\% more success rate (Suc) than model trained only based on MLE. 

\subsection{Case Study}
In this section, we present one randomly sampled example of representative systems and analyze the strengths and weaknesses of them. 
Table~\ref{tab:case_study} shows the example outputs on the \textsc{Yelp} and \textsc{Gyafc} datasets.
We can observe that: (1) The CrossAlign system, which learns a style-independent content  representation vector via adversarial training, tends to sacrifice the content preservation. (2) The Template and Del-Retri systems, which directly removes the specific  style attribute words in the input, can better preserve the content. However, these two systems may fail when the style is implicitly expressed in the input (See the informal-to-formal example). (3) Promisingly, our model achieves a better balance among preserving content, changing the style and improving fluency.

\subsection{Error Analysis}
Although the proposed method outperforms the state-of-the-art systems, we also observe a few failure cases. The typical type of failure cases is that the analogy or metaphor of the style (sentiment). A representative example is ``\textit{over cooked so badly that it was the consistency of canned tuna fish}''. The \textit{canned tuna fish} does not represent its literal content but just an analogy of \textit{``over cooked''}. However, it is really hard for our system as well as other existing methods to balance between preserving the original content and transferring the style when encountering such analogy examples. 

\section{Related Work}
Recently, increasing efforts have been devoted to unsupervised text style transfer.
Despite the increasing efforts devoted to the text style transfer in recent years, the lack of parallel data is still the major challenge for this task.

In order to relief from the need of parallel data, early works generally learn style-independent content representation. In this way, they can train the style rendering model using non-parallel data only.
\cite{fu2017style} leverages the adversarial network to make sure that the content representation does not include style representation. \cite{shen2017style,hu2017toward,yang2018LM} combine Variational Auto-encoder (VAE) with a style discriminator.
Besides, \cite{backTrans2018} strives to get a style-independent content representation through the English-to-French translation models.
However, some recent works~\cite{li2017adversarial,TCLR18TextRewriting} argue that it is often easy to fool the discriminator without actually removing the style information. In other words, the style-independent content representation in latent space may indeed not be able to be achieved in practice, thus causing bad content preservation~\cite{xu2018unpaired}. On the contrary, \cite{li2018delete,zhang2018learning,xu2018unpaired} propose to separate content and style by directly removing the style words.
The former takes advantage of the prior knowledge that style words only localized in corresponding corpora, while the latter skillfully exploits the self-attention mechanism.
However, this explicit separation is not suitable for the text whose style can only be expressed as a whole.

Another way to relief from the need of parallel data is to construct pseudo-parallel data via back-translation, which achieves promising results in unsupervised machine translation~\cite{artetxe2018unsupervised,lample2017unsupervised,lample2018phrase}.
There are also two most recent works~\cite{zhang2018style,TCLR18TextRewriting} directly adopt unsupervised machine translation methods to this task.
However, learning from pseudo-parallel data inevitably accompanies with the data quality problem, thus further influence the control of the preservation of content and accuracy of style. In contrast, we adopt the reinforcement learning algorithm with specifically designed rewards, which directly ensures the two aims of style transfer (Section~\ref{sec:dual}). Meanwhile, the proposed annealing pseudo teacher-forcing algorithm (Section~\ref{sec:highlight}) not only benefits our model from pseudo-parallel data at the beginning of training, but also gradually gets rid of it in the latter stage of training when the model is completely warmed up and is suitable for training mainly based on DualRL.


\section{Conclusion and Future Work}

In this work, we aim at solving text style transfer by learning a direct \textit{one-step} mapping model for the source-to-target style transfer and a dual mapping model for the target-to-source style transfer.
Due to the lack of parallel data, we propose a dual reinforcement learning algorithm DualRL in order to train the two mapping models solely based on the automatically generated supervision signals. In this way, we do not need to do any explicit separation of content and style, which is hard to achieve in practice even with parallel data.
Experimental results on the sentiment transfer and formality transfer datasets show that our model significantly outperforms the previous approaches,  empirically demonstrating the effectiveness of learning two \textit{one-step} mapping models and the proposed DualRL training algorithm.

Although pre-training and annealing pseudo teacher-forcing are effective, they make the training process complicated. Therefore, how to get rid of them and train the generative model purely based on RL from scratch is an interesting direction we would like to pursue.
Moreover, since the proposed architecture DualRL is generic and simple, future work may extend to other unsupervised sequence-to-sequence generation tasks which lack of parallel data. 

\section*{Acknowledgments}
This paper is supported by NSFC project 61751201, 61876004 and 61772040.
The contact authors are Baobao Chang and Zhifang Sui.

\bibliographystyle{named} 
\bibliography{ijcai19}

\nocite{Luo18WSD1}
\nocite{Luo18WSD2}
\nocite{yang2018sgm}
\nocite{liutianyu1}
\nocite{mashuming1}
\nocite{liutianyu2}
\nocite{liutianyu3}

\end{document}